\documentclass[]{article}

\usepackage{times,latexsym}
\usepackage{url}
\usepackage[T1]{fontenc}
\usepackage{amssymb}
\usepackage{pifont}
\usepackage{graphicx}
\usepackage{adjustbox}
\usepackage{appendix}
\usepackage{arxiv}
\usepackage{hyperref}
\usepackage{url,hyperref,lineno,microtype,subcaption}
\usepackage[onehalfspacing]{setspace}
\usepackage{float}

\title{Label-free prediction of fluorescence markers in bovine satellite cells using deep learning}

\usepackage{authblk}

\author[1]{Sania Sinha}
\author[1]{Aarham Wasit}
\author[2]{Won Seob Kim}
\author[2]{Jongkyoo Kim}
\author[3]{Jiyoon Yi\thanks{\texttt{yijiyoon@msu.edu}}}

\affil[1]{Department of Computer Science and Engineering, Michigan State University, East Lansing, MI}
\affil[2]{Department of Animal Science, Michigan State University, East Lansing, MI}
\affil[3]{Department of Biosystems and Agricultural Engineering, Michigan State University, East Lansing, MI}

\date{}

\begin{document}

\newcommand{\xmark}{\ding{55}}%

\maketitle

\begin{abstract}
Assessing the quality of bovine satellite cells (BSCs) is essential for the cultivated meat industry, which aims to address global food sustainability challenges. This study aims to develop a label-free method for predicting fluorescence markers in isolated BSCs using deep learning. We employed a U-Net-based CNN model to predict multiple fluorescence signals from a single bright-field microscopy image of cell culture. Two key biomarkers, DAPI and Pax7, were used to determine the abundance and quality of BSCs. The image pre-processing pipeline included fluorescence denoising to improve prediction performance and consistency. A total of 48 biological replicates were used, with statistical performance metrics such as Pearson correlation coefficient and SSIM employed for model evaluation. The model exhibited better performance with DAPI predictions due to uniform staining. Pax7 predictions were more variable, reflecting biological heterogeneity. Enhanced visualization techniques, including color mapping and image overlay, improved the interpretability of the predictions by providing better contextual and perceptual information. The findings highlight the importance of data pre-processing and demonstrate the potential of deep learning to advance non-invasive, label-free assessment techniques in the cultivated meat industry, paving the way for reliable and actionable AI-driven evaluations.
\end{abstract}

\section{Introduction}

Advancements in technology are crucial for accelerating and automating the assessment of source cell quality in the cultivated meat industry, which aims to address global food sustainability challenges. Bovine satellite cells (BSCs), isolated from animal muscle tissue, are integral to cultured meat production due to their ability to proliferate and differentiate into skeletal muscle cells, driving tissue formation in cultured meat constructs. Ensuring efficient proliferation and differentiation of these cells is essential for producing high-quality cultured meat that can compete with conventional meat products \cite{messmer_2022, stout_2023}. Traditionally, the evaluation of satellite cell quality has relied on techniques such as immunofluorescence microscopy \cite{lee_2021, kong_2023}. While microscopy is a valuable tool for examining cell morphology, it often provides limited contrast and specificity for complex biological samples, necessitating additional fluorescent dyes or antibodies. These processes involve invasive sample preparation and expert annotation of microscopy data. Additionally, variability in cell isolation and culture conditions among different donors can affect the metabolic state and cellular composition, impacting the binding efficiency and specificity of fluorescent stains \cite{kong_2023}. This underscores the need for non-invasive, label-free methods to assess cell quality, especially considering the heterogeneity present during cell proliferation and differentiation.

Artificial intelligence (AI) has recently been leveraged to automate and streamline cellular image analysis. These efforts include deep learning segmentation of subcellular components to reduce the burden of expert annotation \cite{kromp_2021, bilodeau_2022}. Furthermore, predicting fluorescence signals from more cost-effective bright-field microscopy images can minimize the need for invasive staining \cite{christiansen_2018, ounkomol_2018, cheng_2021}. In 2018, Google first introduced \textit{in silico} labeling, a deep learning approach that predicts fluorescence signals from transmitted light z-stack images of unlabeled samples \cite{christiansen_2018}. Additionally, convolutional neural network (CNN) models based on the U-Net architecture \cite{ronneberger_2015} have demonstrated the ability to predict fluorescence signals for individual subcellular components, such as DNA, cell membranes, and mitochondria, directly from transmitted \cite{ounkomol_2018} and reflective \cite{cheng_2021} light microscopy bright-field z-stack images. These studies highlight the potential of AI-enabled image analysis to bridge the gap between traditional and digital techniques, suggesting a promising direction for improving label-free microscopy for cellular assessment. Yet, these applications have predominantly focused on the biomedical sector, where cells are well-characterized and homogeneous (e.g., continuously proliferating human cancer cell lines), unlike the structurally variable BSCs that require advanced methods for precise assessment.

To address the complexity and variability inherent in BSC differentiation, it is crucial to incorporate enhanced visualization techniques into the analysis pipeline. These techniques improve interpretability and explainability, making it easier for researchers to understand model predictions. Recent advancements have demonstrated how improved visualization methods can be applied to biological image analysis, providing clearer insights and improving the reliability of AI predictions \cite{samek_2017}. Applying these techniques in predicting fluorescence or colorimetric signals has shown promise in bridging the gap between traditional and digital techniques \cite{binder_2021, cho_2022}. Therefore, integrating enhanced visualization methods is essential for advancing non-invasive, label-free techniques in the assessment of BSC quality, ultimately supporting the development of reliable and actionable AI-driven assessments.

In this study, we demonstrate a label-free approach for quality assessment of cell culture isolated from bovine muscle tissues, employing deep learning to predict fluorescence signals from bright-field microscopy images. Specifically, we used two key biomarkers to determine the abundance of BSCs in the isolated cell culture, i.e., 4',6-diamidino-2-phenylindole (DAPI) and paired box protein 7 (Pax7). DAPI is a widely used fluorescent stain that binds to cell DNA in fixed cells and tissues, while Pax7 serves as a transcription factor regulating the development and maintenance of skeletal muscle tissue, recognized as the most specific marker of satellite cells \cite{seale_2000, ding_2018}. Consequently, co-staining of cell cultures with DAPI and Pax7 is commonly utilized to monitor the proliferation and differentiation abilities of satellite cells over time \cite{ding_2018, von_2013}. We employed a deep CNN based on the U-Net architecture, adapted from a previous study \cite{ounkomol_2018} with a modified image pre-processing pipeline. The model architecture was trained on our microscopy images to predict multiple fluorescence markers from a single bright-field image of isolated BSCs. Overall, this deep learning approach provides digital staining by learning the features of subcellular components without invasive sample preparation, thereby accelerating cell quality assessment and reducing resource demands.

\section{Methods}

\begin{figure}
    \begin{center}
    \includegraphics[width=1\linewidth]{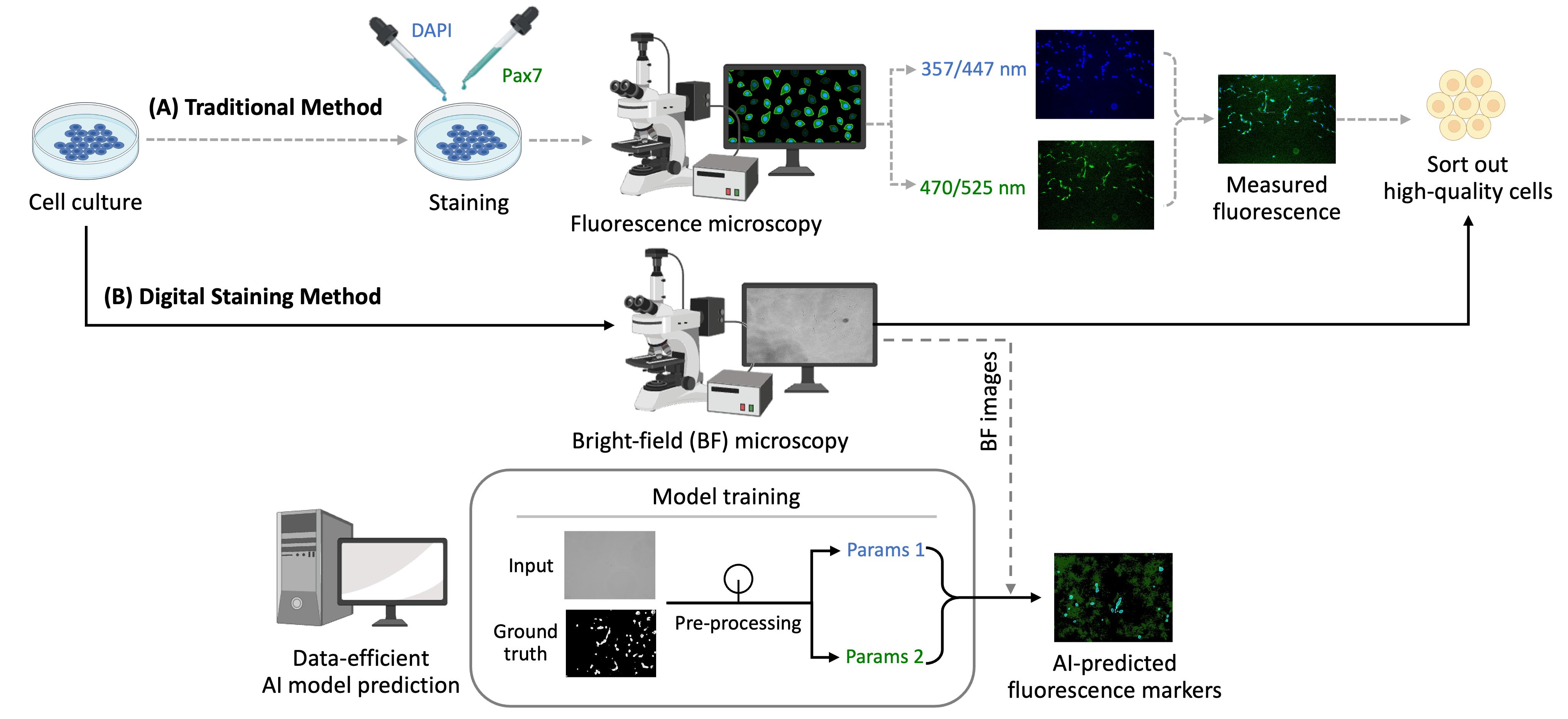}
    \end{center}
    \caption{Schematic diagram for quality assessment of bovine cell culture: (A) Traditional immunofluorescence microscopy method; (B) Digital staining method using deep learning for predicting fluorescence markers. Brightness and contrast of the example images were adjusted for publication clarity.}
    \label{fig:fig1}
\end{figure}

As illustrated in Figure \ref{fig:fig1}, image datasets were obtained using the traditional immunofluorescence microscopy method (Figure \ref{fig:fig1}A). These datasets were used to train the CNN model architecture for digital staining, enabling the trained model to directly predict fluorescence markers in bright-field images without invasive sample preparation (Figure \ref{fig:fig1}B). A set of bright-field (i.e., \textit{input}) and fluorescence (i.e., \textit{ground truth}) images were collected as detailed in Section \ref{sec:data_collection}. The technical details of AI prediction are provided in Section \ref{sec:AI_prediction}, including image pre-processing, model training and prediction, and post-processing.

\subsection{Data collection}
\label{sec:data_collection}

\subsubsection{Cell isolation and culture}

The cell isolation protocol for this project (PROTO202000294) underwent a rigorous review and approval process by the Michigan State University Institutional Animal Care and Use Committee (IACUC). BSCs were meticulously extracted from three-month-old Holstein bull calves (\textit{n} = 3, body weight: 77.10 ± 2.02 kg). These calves were processed with utmost care at the Michigan State University Meat Laboratory (East Lansing, MI, USA) under USDA inspection. The procedures for isolating and cultivating BSCs were meticulously followed, adhering to the methodologies outlined in prior studies \cite{kim_2023a}. Following euthanization using a captive bolt, Longissimus muscle tissue was collected with a sterile knife and immediately transported to the fume hood in PBS with 3× Antibiotic-Antimycotic (Thermo Fisher, Waltham, MA, USA). The muscle tissue was processed to remove blood vessels, connective tissue, and adipose tissue, which were then passed through a sterile meat grinder. The ground muscle was incubated with 0.1\% Pronase® (Calbiochem, La Jolla, CA, USA) and Earl's Balanced Salt Solution (Sigma Aldrich, St. Louis, MO, USA) in a shaking water bath at 37°C for 1 hour. Following incubation, the mixture was centrifuged at room temperature at 1,500 ×g for 4 min. The supernatant was discarded, and the pellet was resuspended in phosphate-buffered saline (PBS; Sigma Aldrich). The resuspended cells were centrifuged at room temperature at 500 ×g for 10 min, and the supernatant was removed. This process was repeated to obtain a pellet of mononucleated cells.

\subsubsection{DAPI and Pax7 staining}

Prior to microscopy imaging, cell culture was stained following the previously published method \cite{kim_2023b}. Briefly, cells were seeded onto 4-well Lab-Tek chamber slides (Thermo Fisher) and incubated in Dulbecco's Modified Eagle's Medium (Gibco, Waltham, MA, USA) supplemented with 10\% Fetal Bovine Serum (Thermo Fisher Scientific) and 1× Antibiotic-Antimycotic (Thermo Fisher) at 38°C in a humidified atmosphere of 95\% O$_{2}$ and 5\% CO$_{2}$ for 24 h. The slides were then fixed with 4\% paraformaldehyde (Thermo Fisher) for 15 min at room temperature. After three PBS washes, cells were permeabilized with 0.1\% Triton X-100 (Thermo Fisher) in PBS for 15 min. To block non-specific binding, a 2\% bovine serum albumin (Thermo Fisher) solution in PBS was applied and incubated for 1 h at 4°C. Cells were incubated overnight at 4°C with primary antibodies: anti-Pax7 (mouse monoclonal, 1:500, Developmental Studies Hybridoma Bank, Iowa City, IA, USA). Following three PBS washes, cells were incubated with Alexa Fluor 488 anti-mouse IgG secondary antibody (1:1000; Thermo Fisher) for 30 min at room temperature. After three more PBS washes, cells were counterstained with DAPI (1:1000; Thermo Fisher) in PBS for 5 min at room temperature. Coverslips were mounted onto glass slides using Fluoromount-G™ Mounting Medium (Thermo Fisher) and sealed with nail polish. 

\subsubsection{Bright-field and fluorescence microscopy}
\label{sec:microscopy}

The stained slides were imaged using an inverted digital microscope (EVOS M5000, Thermo Fisher) at 20× magnification in three different modes: bright-field, DAPI fluorescence (excitation/emission: 357/447 nm), and green fluorescent protein fluorescence (470/525 nm). The images were obtained using optimal brightness parameters (light intensity, exposure time, and gain) for each mode: bright-field with 18.88\% intensity, 16 ms exposure, and 1 dB gain; DAPI with 7.553\% intensity, 52.2 ms exposure, and 30.6 dB gain; and Pax7 with 43.75\% intensity, 94.4 ms exposure time, and 114 dB gain. Background fluorescence scans were acquired using blank chamber slides for the DAPI and GFP fluorescence ranges. Triplicate sets of bright-field and fluorescence (both DAPI and Pax7) microscopy images were collected per biological sample, and 48 biological replicates were used.

\subsection{AI prediction of fluorescence markers}
\label{sec:AI_prediction}

\subsubsection{Image pre-processing pipeline}

\begin{figure}[h!]
    \begin{center}
    \includegraphics[width=1\linewidth]{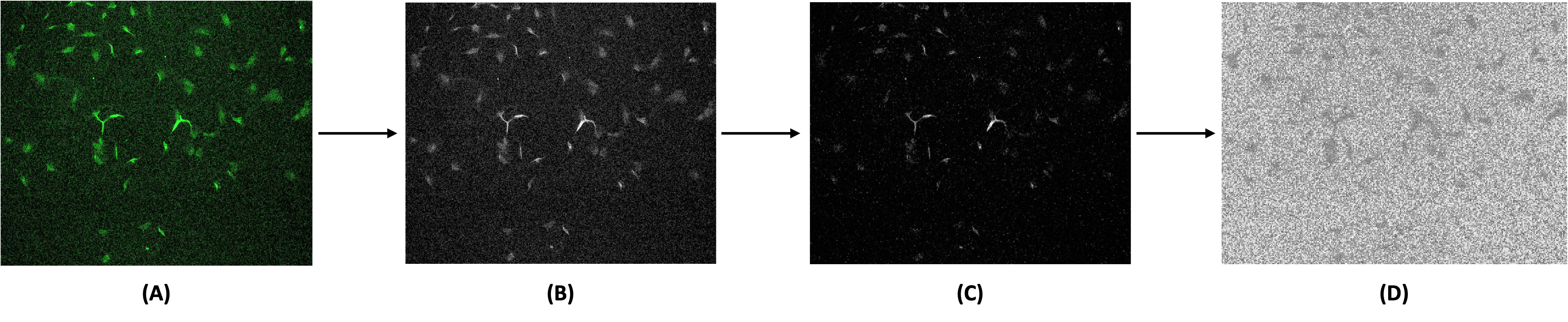}
    \end{center}
    \caption{Image pre-processing pipeline for generating \textit{target} fluorescence signals from \textit{ground truth} fluorescence images. (A) Original \textit{ground truth} data obtained by fluorescence microscopy were converted into (B) grayscale images, followed by (C) fluorescence denoising, and then (D) final normalization to enhance the \textit{target} fluorescence signals for model training. Brightness and contrast of the example images were adjusted for publication clarity.}
    \label{fig:fig2}
\end{figure}

The collected \textit{ground truth} data consisted of single-channel, colored fluorescence images. These images were pre-processed to generate \textit{target} fluorescence signals, as shown in Figure \ref{fig:fig2}. Since the CNN architecture used supports only grayscale images, the raw TIF images were first converted into grayscale. Next, background noise was reduced by subtracting the average of the background fluorescence scans during the fluorescence denoising step. This process helped remove random bright spots and normalize areas of high brightness, ensuring uniform identification of subcellular structures. Standard averaging techniques were ineffective due to the randomness of noise and the placement of subcellular components. The processed images were then split into \textit{training} and \textit{testing} datasets. 

\subsubsection{Model training and prediction}

The PyTorch-fnet framework originally developed for human cells by \cite{ounkomol_2018}, a CNN model based on the U-Net architecture for label-free fluorescence prediction, was employed in this study. Built on the PyTorch library \cite{paszke_2019}, the fnet model includes various options for transformations and metrics, which can be selected during the training parameter setup through a JSON file, instead of having to modify the model itself. The fnet model offers neural networks for both single-channel and multi-channel data. For this study, a PyTorch normalize transform was used, and the \textit{fnet\_nn\_2d} model architecture was selected. The loss function employed was a weighted mean squared error. A CSV file containing the paths of the bright-field images with their respective fluorescence counterparts for the \textit{training} dataset was fed into the model. Following training, the model predicted fluorescence signals on an unseen \textit{testing} dataset. Prediction parameters were set using a JSON file to ensure a consistent normalize transform with training, with the correlation coefficient as an in-built evaluation metric.

\subsubsection{Post-processing of model prediction results}

To enhance the interpretability of the AI model predictions, post-processing steps were employed. While the original \textit{ground truth} data were colored, the model was designed to use grayscale bright-field images as input and produce grayscale outputs, following the approach by \cite{ounkomol_2018}. These model prediction outputs, initially in TIF format, were converted from grayscale to RGB and then to JPG format for visualization. Post-processing also involved color mapping, a standard digital image enhancement technique \cite{faridul_2014}. The original \textit{ground truth} data were used to devise a color palette that maps the colorized output images closest to the original image selection. This step made the \textit{predicted} images more consistent with traditional \textit{ground truth} data and enhanced the visibility of subcellular components, particularly for noisy outputs.

The final step involved merging the output predictions for DAPI and Pax7 markers to produce the desired result of combined fluorescence markers. This was accomplished using scripts for image overlay and transparency adjustment, resulting in a more accurate prediction of the location and density of satellite cells. By merging the colorized predictions, we created a comprehensive visualization that resembles traditional multi-channel fluorescence microscopy.

\subsubsection{Model performance evaluation}

Evaluating a model is essential for determining its effectiveness. However, establishing evaluation metrics or error values that accurately reflect model performance can be challenging. To address this, multiple standard statistical performance metrics were employed to assess model performance from various perspectives.

The Pearson correlation coefficient \cite{nettleton_2014}, also used in the study by \cite{ounkomol_2018}, measures the normalized covariance between the \textit{target} and \textit{predicted} images, with values ranging from -1 to 1. Values closer to 1 indicate higher correlation and image similarity. Mathematically, the absolute value of the Pearson correlation coefficient is given by:

$$
r = \Bigg |\frac{\sum (x_i - \bar{x})(y_i - \bar{y})}{\sqrt{\sum(x_i - \bar{x})^2\sum(y_i - \bar{y})^2}} \Bigg |
$$

where $x_i$ and $y_i$ are the individual data points, and $\bar{x}$ and $\bar{y}$ are the respective means.

In addition, other widely used metrics such as Mean Squared Error (MSE) and Structural Similarity Index (SSIM) \cite{wang_2004} were calculated. MSE, one of the most general measures of error, was computed by taking the average squared difference between the pixels of the \textit{target} and \textit{predicted} images. SSIM considers image texture and granularity, providing a more refined measure than simple MSE. Mathematically, the absolute value of SSIM is given by:

$$
SSIM(x,y) =  |[l(x,y)]^a \times [c(x,y)]^\beta \times [s(x,y)]^\gamma|
$$

where
$$
l(x,y) = \frac{2\mu_x\mu_y + C_1}{\mu_x^2 + \mu_y^2 + C_1}
$$
$$
c(x,y) = \frac{2\sigma_x\sigma_y + C_2}{\sigma_x^2 + \sigma_y^2 + C_2}
$$
$$
s(x,y) = \frac{\sigma_{xy} + C_3}{\sigma_x\sigma_y + C_3}
$$

Here, $\mu_x$ and $\mu_y$ are the pixel sample means, $\sigma_x$ and $\sigma_y$ are the standard deviations, $\sigma_{xy}$ is the covariance, and $C_1$, $C_2$, and $C_3$ are constants to stabilize the division with weak denominators.

\section{Results and Discussion}

\subsection{Evaluation of model performance with enhanced visual interpretability}

To evaluate model performance, the \textit{predicted} fluorescence images were qualitatively compared to the \textit{target} fluorescence signals. The use of post-processing techniques, including color mapping and image overlay, facilitated a clearer interpretation of the fluorescence signals. These techniques provided vital contextual information, enhancing the perceptual quality of the predictions. The merged predictions of DAPI and Pax7 markers enabled precise localization of BSCs on input bright-field images, demonstrating the model's capability in digital staining for cell culture quality assessment.

\subsubsection{DAPI predictions exhibit better performance compared to Pax7}

\begin{figure}[h!]
    \begin{center}
    \includegraphics[width=1\linewidth]{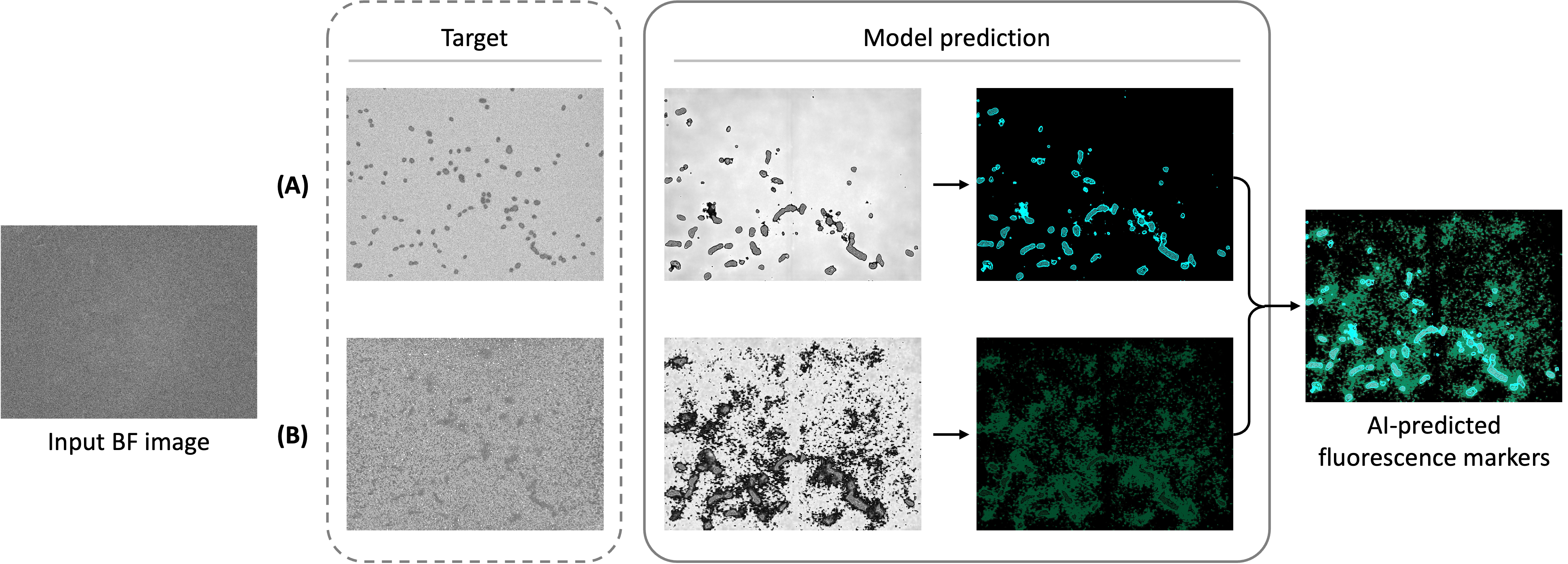}
    \end{center}
    \caption{Example images of model prediction of multiple fluorescence markers from a single bright-field (BF) image: (A) DAPI and (B) Pax7. Model predictions include direct output followed by post-processing using digital image enhancement techniques.}
    \label{fig:fig3}
\end{figure}

As shown in Figure \ref{fig:fig3}, the model predictions for DAPI achieved better performance compared to Pax7. The DAPI predictions displayed less background noise and variability, attributed to the uniform staining and distribution of DAPI, which binds to DNA. In contrast, Pax7 predictions were more variable due to the inconsistent expression and localization of Pax7 in cells. This observation suggests that the fnet model architecture is particularly well-suited for predicting DAPI fluorescence, aligning with its original design for subcellular structures like DNA and cell membranes \cite{ounkomol_2018}.

\subsubsection{Biological implications in improving model performance}

The use of DAPI and Pax7 in this study was intended to assess the proliferation and differentiation capabilities of BSCs. These fluorophores target specific cellular components, enhancing contrast and resolution. However, biological samples often exhibit noisy backgrounds and diffuse signals, particularly with Pax7, due to the heterogeneity in myogenic differentiation of BSCs \cite{kong_2023}. This variability poses significant challenges for signal quantification and automated analysis.

Deep learning techniques rely heavily on high-quality data and tend to underperform when such data are scarce. This issue is particularly relevant in predicting immunofluorescence signals like Pax7, where the limited availability of labeled data exacerbates the challenge. Rather than training a model from scratch, fine-tuning a pre-trained model with local data has been shown to be a more effective strategy \cite{tajbakhsh_2016, moen_2019}. Thus, future studies should focus on improving pre-training strategies specifically for Pax7 with heterogeneous biological states. This could involve using attention-based networks to segment subcellular components with varying health states \cite{wang_2023}, or incorporating deep learning-based identification of cell differentiation \cite{zhu_2021}. Enhancing the handling of Pax7 signals is crucial for advancing the reliability of deep learning models in predicting these markers.

\subsection{Improved consistency of predictions through fluorescence denoising}

\begin{figure}[h!]
    \begin{center}
    \includegraphics[width=0.9\linewidth]{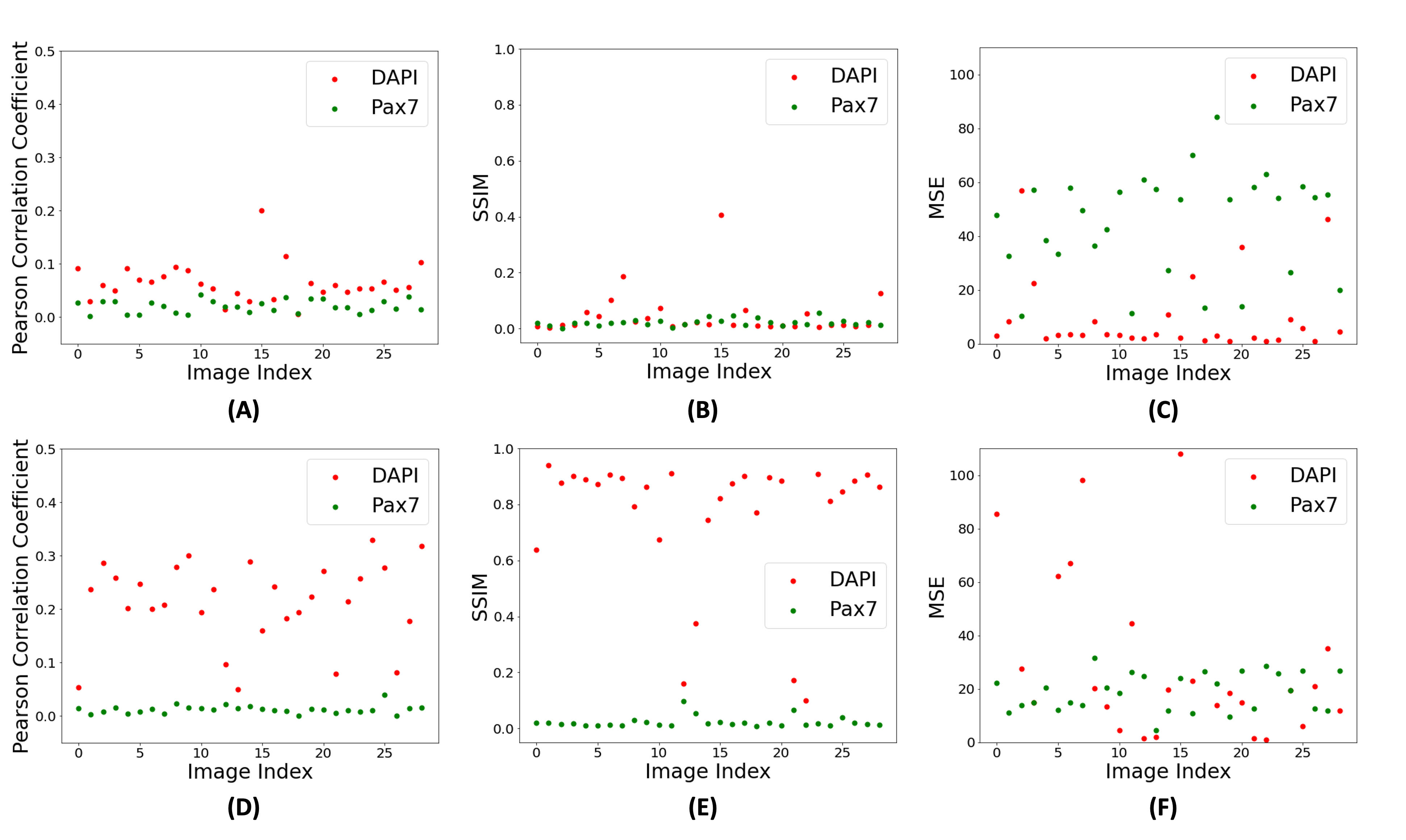}
    \end{center}
    \caption{Statistical performance evaluation metrics for DAPI and Pax7 predictions without (top row) and with (bottom row) fluorescence denoising. (A,D) Pearson correlation coefficient: higher values indicate better correlation. (B,E) SSIM: Structural Similarity Index, higher values indicate greater similarity. (C,F) MSE: Mean Squared Error, lower values indicate better accuracy.} 
    \label{fig:fig4}
\end{figure}

In addition to the individual visual assessment of model performance, its consistency was investigated using selected statistical performance evaluation metrics: Pearson correlation coefficient, SSIM, and MSE. The top row of Figure \ref{fig:fig4} illustrates the performance of the model trained on \textit{target} fluorescence signals derived from raw data without fluorescence denoising. All three metrics showed similar trends for both DAPI and Pax7 predictions, with slightly higher values for DAPI predictions in the Pearson correlation coefficient. The average Pearson correlation coefficient for DAPI was 0.0649, SSIM was 0.047, and MSE was 9.507. For Pax7, the average Pearson correlation coefficient was 0.020, SSIM was 0.022, and MSE was 44.753. The bottom row Figure \ref{fig:fig4} shows the changes in these evaluation metric values after applying our image pre-processing pipeline for fluorescence denoising, as depicted in Figure \ref{fig:fig2}. This pre-processing resulted in higher values for SSIM and Pearson Correlation Coefficient metrics, indicating an overall improvement in model performance. For MSE, we got lower values for Pax7 indicating an improvement, but values for DAPI increased. After denoising, the average Pearson correlation coefficient for DAPI increased to 0.212, SSIM to 0.761, and MSE to 41.571. For Pax7, the average Pearson correlation coefficient increased to 0.124, SSIM to 0.023, while MSE decreased to 18.793.

\subsubsection{Feasibility of evaluation metrics in digital staining}

As shown in Figure \ref{fig:fig4}, MSE values increased with fluorescence denoising in our image pre-processing pipeline, suggesting greater errors in pixel-wise predictions. Despite this, visual assessment of the final outputs showed improved model performance with our image pre-processing. This improvement is consistent with the increase in the Pearson correlation coefficient and SSIM values, indicating better correlation and structural similarity between the \textit{target} and \textit{predicted} signals. This discrepancy suggests that MSE may not be the most appropriate metric for evaluating model performance in this context. While MSE measures pixel-wise accuracy, it may not fully capture perceptual quality, spatial context, or signal-to-noise ratio. Perceptual quality, relating to human visual perception, is better captured by metrics like SSIM that consider structural information \cite{wang_2004}. SSIM evaluates luminance, contrast, and structure, making it more sensitive to visual perception than MSE. Spatial context is crucial in biological imaging, where the arrangement and relationship of cellular structures matter more than exact pixel values. SSIM captures spatial information and provides a better understanding of image quality \cite{zhou_2002}. Furthermore, the signal-to-noise ratio is critical in microscopy images, where high background noise can obscure meaningful signals. MSE does not account for noise distribution, whereas SSIM can provide a more nuanced assessment of image quality by considering noise levels and their impact on structural similarity \cite{brunet_2012}. Overall, while MSE measures pixel-wise accuracy, it falls short in capturing the perceptual quality, spatial context, and signal-to-noise ratio essential for evaluating digital staining in cell microscopy.

Additionally, the Pearson correlation coefficient measures the linear relationship between \textit{target} and \textit{predicted} signals, providing insights into overall trend alignment rather than pixel-wise accuracy. The average Pearson correlation coefficient obtained in this study was lower than in the original study of the fnet model \cite{ounkomol_2018}, where the value for DNA was over 0.6. This discrepancy can be attributed to differences in the input data. The previous study used 3D z-stacks of bright-field images or 2D electron micrographs, which provide more comprehensive information about subcellular structures and thus achieved higher correlation values. In contrast, our study used only single focal plane data, which may lack some spatial context. Despite this, our model still performed reasonably well, demonstrating the robustness of our approach in predicting fluorescence signals from 2D bright-field data. This adaptation underscores the practical applicability of deep learning in image-based quality assessment of BSC culture, enabling cost-effective digital staining in cell imaging.

\subsubsection{Importance of pre-processing in addressing biological heterogeneity}

To effectively utilize existing models and fine-tune them to specific datasets, data pre-processing is essential, especially in managing inconsistent data quality and mitigating the risk of overfitting. In our approach to fluorescence denoising, background fluorescence scans were subtracted from the raw data. Standard normalization techniques were ineffective due to randomly scattered noise, which often removed the actual areas of interest. This noisy fluorescence background in the raw data necessitated optimization of brightness parameters, such as light intensity, exposure time, and gain for each fluorescence channel, as described in Section \ref{sec:microscopy}, resulting in inadvertently elevated nonspecific background fluorescence. To address this issue, a fluorescence denoising technique for each channel was implemented in our pre-processing pipeline (Figure \ref{fig:fig2}), which substantially improved the consistency of model predictions, as demonstrated in Figure \ref{fig:fig4}. This adjustment enhances the reliability of results by accommodating variability in brightness parameters across different fluorescence channels. Moreover, researchers have explored various experimental approaches to improve staining methods and reduce nonspecific binding \cite{zaqout_2020}. Additionally, algorithms have been developed to digitally remove autofluorescence signals \cite{wang_2022}. These efforts underscore the ongoing need to improve fluorescence specificity in the quantitative assessment of microscopy images. Continued research is essential to enhance the quality of \textit{training} data, thereby advancing the application of deep learning for precise fluorescence prediction in cell imaging.

\section{Conclusions}

In summary, our study demonstrates a label-free approach for assessing BSC cultures using deep learning to predict multiple fluorescence signals from a single bright-field microscopy images. By utilizing DAPI and Pax7 as biomarkers, and employing a U-Net-based CNN model with an enhanced pre-processing pipeline, we significantly improved prediction performance and consistency. The model performance was effectively evaluated using Pearson correlation coefficient and SSIM, which better captured perceptual quality and spatial context than traditional MSE. Enhanced visualization techniques provided crucial contextual information, improving the interpretability of predictions. Our findings highlight the importance of data pre-processing and underscore the potential of deep learning to advance non-invasive, label-free assessment techniques in the cultivated meat industry.

\section*{Acknowledgements}

This work was supported by the Michigan State University startup funds for JY.

\bibliographystyle{plain}

\end{document}